**Validity of a clinical decision rule–based alert system for drug dose adjustment in patients**

**with renal failure intended to improve pharmacists' analysis of medication orders in**

**hospitals**

Boussadi A[1,2,4], Caruba T[2,3], Karras A[2], Berdot S[1,2,4], Degoulet P[1,2], Durieux P[1,2], Sabatier B[1,2]

1. Paris Descartes University (Paris 5), Paris, France and INSERM UMR_S 872 Eq 22

2. Assistance Publique – Hôpitaux de Paris, Hôpital Européen Georges Pompidou, Paris, France

3. INSERM U 756 and LIRAES EA 4470, Paris, France

4. UPMC University (Paris 06), Paris, France

*Errors/prevention & control; Pharmaceutical Preparations/administration & dosage; Drug*

*Dosage Calculations ; Drug Prescriptions; Drug Therapy;*

***Word count***

*Body: 3753*

*Abstract: 280*

**Correspondence:** Abdelali Boussadi, *Dipl.-Ing.*, *M.Sc.*, *PhD*, Département d'Informatique Hospitalière (DIH)-Hôpital Européen Georges Pompidou, 20 Rue Leblanc, 75908 Paris Cedex 15, France; e-mail: <abdelali.boussadi@egp.aphp.fr>.

**Abstract**

**Objective:** The main objective of this study was to assess the diagnostic performances of an alert system integrated into the CPOE/EMR system for renally cleared drug dosing control. The generated alerts were compared with the daily routine practice of pharmacists as part of the analysis of medication orders. **Materials and Methods:** The pharmacists performed their analysis of medication orders as usual and were not aware of the alert system interventions that were not displayed for the purpose of the study neither to the physician nor to the pharmacist but kept with associate recommendations in a log file. A senior pharmacist analyzed the results of medication order analysis with and without the alert system. The unit of analysis was the drug prescription line. The primary study endpoints were the detection of drug dose prescription errors and inter-rater reliability between the alert system and the pharmacists in the detection of drug dose error. **Results:** The alert system fired alerts in 8.41% (421/5006) of cases: 5.65% (283/5006) "exceeds max daily dose" alerts and 2.76% (138/5006) "under-dose" alerts. The alert system and the pharmacists showed a relatively poor concordance: 0.106 (CI 95% [0.068 – 0.144]). According to the senior pharmacist review, the alert system fired more appropriate alerts than pharmacists, and made fewer errors than pharmacists in analyzing drug dose prescriptions: 143 for the alert system and 261 for the pharmacists. Unlike the alert system, most diagnostic errors made by the pharmacists were 'false negatives'. The pharmacists were not able to analyze a significant number (2097; 25.42%) of drug prescription lines because understaffing. **Conclusion:** This study strongly suggests that an alert system would be complementary to the pharmacists' activity and contribute to drug prescription safety.

# 1. Introduction

Clinical decision support systems (CDSS) have already proven their efficacy in preventing medication errors related to prescribing (1)(2). However, studies evaluating computerized physician order entry systems (CPOE) and CDSS have also reported a reluctance of physicians to use computerized tools. Pharmacists may be more willing to use computerized tools than physicians (3). Physicians frequently ignore justified CDSS medication alerts whereas pharmacists help physicians handle these alerts correctly (4). Pharmacists are also effective at preventing adverse drug events (5)(6) and reducing costs associated with drug therapies (7).

In France, pharmaceutical analysis of medication orders has been mandatory since 1991 in hospitals (8). Pharmacists check physicians' medication orders for each patient to detect possible errors. The daily analysis of all the orders makes up a large proportion of the workload of pharmacists. The medication orders associated most commonly with errors are those for cases involving decline in renal or hepatic function requiring modifications of drug therapy (9).

A CDSS adapted to this activity could potentially improve the safety of medication prescriptions at a reasonable cost without imposing an unacceptable workload on pharmacists.

We designed and implemented a clinical decision rule–based alert system for drug dose verification integrated into the CPOE system of the Hôpital Européen Georges Pompidou (HEGP) (10)(11). The alert system was designed to check medication orders and, if necessary, display recommendations to adjust medication doses for patients suffering from impaired renal function. The main objective of this study was to assess the diagnostic performances of the alert system compared with the routine daily practice of pharmacists concerning the pharmaceutical analysis of medication orders.

## 2. Methods

### 2.1 Study site and settings

HEGP is a teaching hospital with 24 clinical departments and 795 beds. The HEGP clinical information system integrates an electronic health record (EHR) with a CPOE (DxCare®, MEDASYS) (12). In DxCare®, each medication order includes one or several 'prescription lines'. Each prescription line corresponds to one prescribed drug.

At HEGP, the pharmaceutical analysis of medication orders is part of the daily practice for medication orders in nine clinical departments containing 227 beds: nephrology (19 beds), vascular medicine (23 beds), clinical immunology (19 beds), cardiovascular surgery (52 beds), geriatric health service (27 beds), orthopedics (17 beds), cardiology (29 beds), hypertension (16 beds) and internal medicine (25 beds). The mean time of hospital stay in these departments is 6 days (range 3 to 11 days). The mean number of medication orders is nine per day per patient. The team of pharmacists responsible for the daily pharmaceutical analysis of medication orders is composed of six pharmacists (three seniors and three residents). Every day, each pharmacist accompanies physicians on their rounds and analyses medication prescription orders from one or two clinical departments. Each medication order is checked, drug line by drug line (dose, unit, time of administration, route of administration, frequency per day, reconstitution process, and compatibility) against biological markers and patient records.

Although pharmaceutical analysis of medication orders has been mandatory in hospitals in France since 1991 (8), there are insufficient pharmacists for this task so not all physicians' medication orders are validated.

This study was conducted over a period of 7 months between March 2011 and September 2011. All patients admitted in one of the nine departments with pharmaceutical analysis activities and prescribed one or more medications targeted by the alert system (Table I) were included in this study.

## 2.3 Clinical alert system design

Using the Summary of Product Characteristics, the recommendations of the "drug prescribing in renal failure: dosing guidelines for adults and children" (13) and the French Guidelines for Prescription in Renal Disease (GPR) handbook (14), an expert panel, including nephrologists and pharmacists, determined the dose adjustments according to the estimated glomerular filtration rate (eGFR) for 24 drugs (Table I). The eGFR was estimated using the revised-4-component version of the Modification of Diet in Renal Disease (MDRD) study equation; however, the race component was not taken into account because it has not been validated in the European context. Required dosing adjustments were based on degree of kidney function impairment, classified into different categories according to the medication. For example, for Allopurinol, four categories were defined: eGFR (< 15 mL/min/1.73 m2), eGFR (between 15 to 30 mL/min/1.73 m2), eGFR (between 30 to 59 mL/min/1.73 m2) and eGFR (> 59 mL/min/1.73 m2). The expert panel determined the dose range and frequency of administration for each of the medications in each of these categories and described them in a set of recommendation tables (15).

Using a clinical decision rule design framework called BRDF, described in (11), we established 962 clinical decision rules (Table V) to fire "exceeds max daily dose" alerts and "under-dose" alerts according to the recommendation tables (15). We implemented these rules as an alert system using a Business Rule Management System (BRMS) from IBM® which allows the pharmacists to be included in the implementation step through the 'user friendly' interface called Rule Team Server (RTS). To refine the decision rules implemented and before the integration of the alert system into the CIS, several iterations of a retrospective validation step were run using a clinical data warehouse (CDW). This design step is described in (15). Finally we integrated the alert system to the CPOE/EMR system of the HEGP.

## 2.4 Study design

### The pharmacists' daily analysis of the medication orders

A study at the HEGP to assess the individual variation between pharmacists analysis of physicians' medication orders revealed substantial disparity between pharmacists (manuscript in preparation). Thus, it was important to standardize the pharmaceutical analysis activities between the pharmacists to facilitate comparisons with the alert system analysis. We therefore organized several training sessions for the pharmacists to be familiar with the recommendation tables described previously.

During the study period, the pharmacists performed the analysis of the medication orders as usual; if a drug prescription line contains an error, they report it using the CPOE interface dedicated to the pharmacists. Otherwise, the pharmacist tags the drug prescription line as 'accepted'. For the purpose of this prospective study the alerts and recommendations targeted to the physicians and/or pharmacists were not displayed in the CPOE interface but stored in a log file to be compared later with the pharmacists' analysis of medication orders. The panel of pharmacists was therefore not aware of the alert system analysis of the medication orders.

### The alert system analysis of the medication orders

During the study period, the alert system automatically checked drug doses in medication orders including one or several drugs listed in Table I and fired an alert with a recommendation if necessary. Otherwise, the alert system tagged the drug prescription line as 'accepted'. Each week, the alert system stored all the results in a CSV log file. Each line of the log file corresponds to a CPOE-drug prescription line and includes the following data: Patient identifier, encounter number, patient first and last name, patient date of birth, patient sex, hospitalization department, medication, dose regimen, prescription start and end date, patient eGFR, pharmacist recommendation and alert system recommendation, and rule ID.

**Impact on clinical decision rules**

During the study period, each alert fired by the alert system was tagged with a rule ID. Consequently, each drug prescription line reviewed by the senior pharmacist and liable to an alert was also related to a rule ID. We used this relation to assess the utility of this study for the rule refinement and to make some improvements.

**Statistical analysis**

The unit of analysis was the drug prescription line; all the drug prescription lines analyzed both by the alert system and the pharmacists were studied. The primary study endpoints were the detection of drug dose prescription errors and the inter-rater reliability between detection of drug dose errors by the alert system and by the pharmacists. A drug prescription line is considered erroneous if the drug dose is higher (max daily dose error) or lower (under dose error) than the drug dose recommended by the expert panel according to the patient eGFR (15). Agreement between the pharmacists and the alert system was assessed using Cohen's Kappa coefficient. Its values range from -1 to 1 and the closer the coefficient to one, the higher the level of agreement between the alert system and the pharmacists.

At the end of this step, a senior pharmacist analyzed all discordances between the alert system and the pharmacists and rated the potential severity of each drug prescription identified as erroneous by the alert system or by the pharmacists according to a three-category scale used in other publications (16): none; purely preventive; and serious or significant and life-threatening. The same senior pharmacist analyzed a random sample of concordant medication orders to estimate the concordance rate of alerts fired by the alert system and the pharmacists.

## 4. Results

A total of 3228 patients were included in the study, and six pharmacists were involved (three seniors and three residents). The number of prescriptions analyzed per pharmacist during the study was 903 (median IQR [368-1132]).

During the study period, 29 log files were generated by the alert system. These files included 8251 drug prescription lines, including 1148 (13.91%) with no specified dose because they were 'conditional' prescription lines, and 2097 (25.42%) were not analyzed by a pharmacist because of understaffing; these prescription lines were excluded from the subsequent analyses. In fine, 5006 (60.67%) drug prescription lines had both pharmacist and alert system recommendations (Figure 1), involving 27 different medications (Table I).

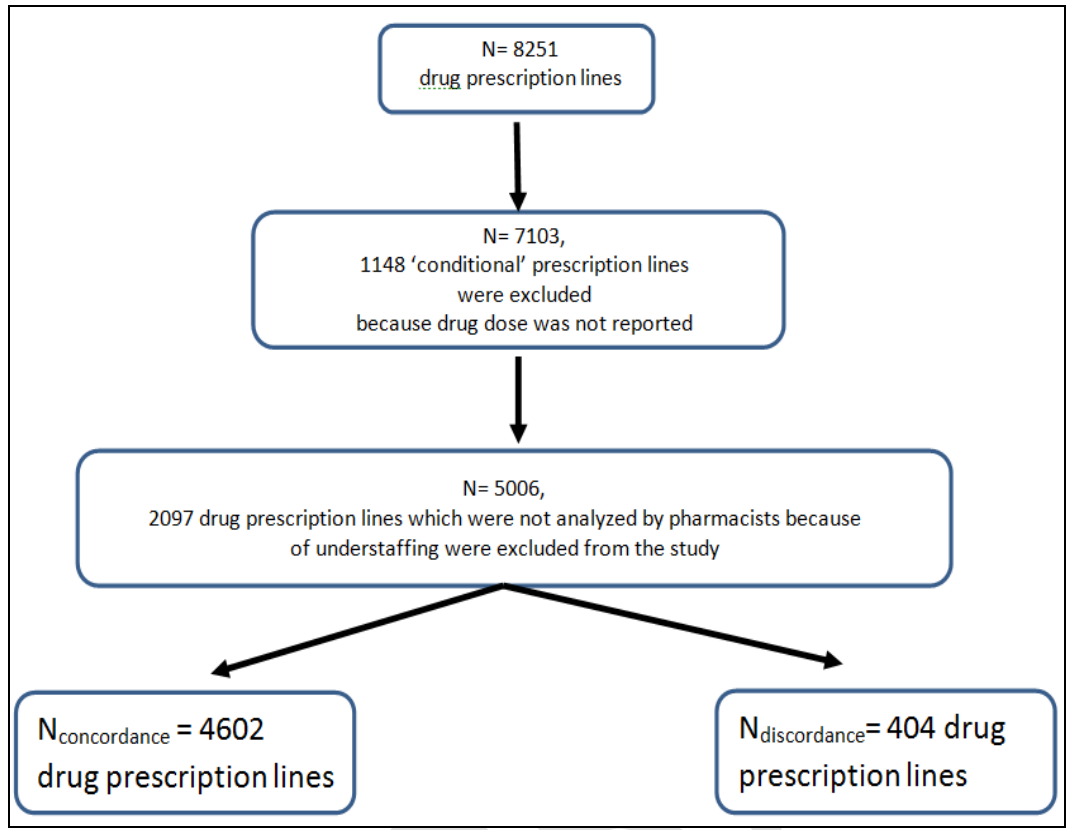

**Figure 1:** study flowchart showing the number of drug prescription lines included/excluded from the study.

**Table I:** Frequency of alerts fired by the alert system and the pharmacists

| | number of prescription lines | Alert system | | Pharmacists | |
|---|---|---|---|---|---|
| | | Max daily dose alerts frequency | Under-dose alerts frequency | Max daily dose alerts frequency | Under-dose alerts frequency |
| Allopurinol | 309 | 49 (15.86%) | 2 (0.65%) | 6 (1.94%) | 0 (0.00%) |
| Amikacin | 21 | 9 (42.86%) | 2 (9.52%) | 0 (0.00%) | 0 (0.00%) |
| Amorolfine[1] | 2 | 0 (0.00%) | 0 (0.00%) | 0 (0.00%) | 0 (0.00%) |
| Amoxicillin | 310 | 8 (2.58%) | 1 (0.32%) | 1 (0.32%) | 0 (0.00%) |
| Amoxicillin and potassium clavulanate | 440 | 69 (15.68%) | 5 (1.14%) | 2 (0.45%) | 0 (0.00%) |
| Atenolol | 303 | 5 (1.65%) | 8 (2.64%) | 3 (1%) | 0 (0.00%) |
| Bisoprolol | 508 | 0 (0.00%) | 0 (0.00%) | 0 (0.00%) | 0 (0.00%) |
| Captopril | 25 | 0 (0.00%) | 7 (28.00%) | 0 (0.00%) | 0 (0.00%) |
| Cefotaxim | 355 | 2 (0.56%) | 10 (2.82%) | 1 (0.28%) | 1 (0.28%) |
| Ciprofloxacin | 99 | 2 (2.02%) | 2 (2.02%) | 0 (0.00%) | 0 (0.00%) |
| Erythromycin | 3 | 0 (0.00%) | 0 (0.00%) | 0 (0.00%) | 0 (0.00%) |
| Ethambutol | 35 | 7 (20.00%) | 1 (2.86%) | 0 (0.00%) | 0 (0.00%) |
| Fosfomycin | 1 | 0 (0.00%) | 0 (0.00%) | 0 (0.00%) | 0 (0.00%) |
| Gentamicin | 56 | 30 (53.57%) | 6 (10.71%) | 3 (5.36%) | 0 (0.00%) |
| Isoniazid | 38 | 0 (0.00%) | 34 (89.47%) | 0 (0.00%) | 0 (0.00%) |
| Levofloxacin | 230 | 12 (5.22%) | 3 (1.30%) | 4 (1.74%) | 0 (0.00%) |
| Metformin | 387 | 69 (17.83%) | 5 (1.29%) | 10 (2.58%) | 0 (0.00%) |

---

[1] Paraffin-Vaseline and Amorolfine were not in the targeted medications listed in (Table I); however, prescription lines of these medications were reported in the log file by the alert system. The explanation lies in the implementation of the data access layer of the alert system. This layer allows querying the CPOE/EHR database to gather the data that should be checked by the clinical decision rules. In this SQL query, we inadvertently added Paraffine-Vaseline and Amorolfine as targeted medication to be checked by the alert system.

| | | | | | |
|---|---|---|---|---|---|
| **Metronidazole** | 99 | 3 (3.03%) | 0 (0.00%) | 1 (%) | 0 (0.00%) |
| **Norfloxacin** | 19 | 1 (5.26%) | 9 (47.37%) | 0 (0.00%) | 0 (0.00%) |
| **Paraffin, Vaseline**[2] | 110 | 0 (0.00%) | 0 (0.00%) | 0 (0.00%) | 0 (0.00%) |
| **Ramipril** | 1148 | 6 (0.52%) | 16 (1.39%) | 1 (0.09%) | 1 (0.09%) |
| **Sulfamethoxazole, Trimethoprim** | 256 | 0 (0.00%) | 0 (0.00%) | 0 (0.00%) | 0 (0.00%) |
| **Tramadol** | 82 | 6 (7.32%) | 2 (2.44%) | 1 (1.22%) | 0 (0.00%) |
| **Tobramycin** | 2 | 0 (0.00%) | 2 (100.00%) | 0 (0.00%) | 0 (0.00%) |
| **Valaciclovir** | 93 | 0 (0.00%) | 0 (0.00%) | 0 (0.00%) | 0 (0.00%) |
| **Vancomycin** | 62 | 5 (8.06%) | 23 (37.10%) | 1 (1.61%) | 1 (1.61%) |
| | 5006 | 283 (5.56%) | 138 (2.76%) | 34 (0.68%) | 3 (0.06%) |

---

[2] Paraffin-Vaseline and Amorolfine were not in the targeted medications listed in (Table I); however, prescription lines of these medications were reported in the log file by the alert system. The explanation lies in the implementation of the data access layer of the alert system. This layer allows querying the CPOE/EHR database to gather the data that should be checked by the clinical decision rules. In this SQL query, we inadvertently added Paraffine-Vaseline and Amorolfine as targeted medication to be checked by the alert system.

**Alert system performance indicators, comparison with pharmacists' daily analysis of medication orders**

The alert system fired alerts for 8.41% (421/5006) of the lines, including 5.65% (283/5006) "exceeds max daily dose" alerts and 2.76% (138/5006) "under-dose" alerts (Table 1).  Prescriptions of gentamicin had the highest frequency of exceeds max daily dose alerts fired (53.57%) and prescriptions of isoniazid had the highest frequency of under-dose alerts (89.47%).  The alert system did not fire any alert for the prescription lines of seven medications:  Amorolfine, Bisoprolol, Erythromycin, Fosfomycin, Paraffin-Vaseline, Sulfamethoxazole-Trimethoprim, and Valaciclovir. Paraffin-Vaseline and Amorolfine were not in the targeted medications listed in (Table I); however, prescription lines of these medications were reported in the log file by the alert system (Table I).

The pharmacists fired alerts for 0.74% (37/5006) lines; including 0.68% (34/5006) exceeds max daily dose alerts and 0.06% (3/5006) under-dose alerts (Table I).   Prescriptions of metformin had the highest frequency of exceeds max daily dose alerts (2.58%). Pharmacists didn't fire any alerts for 14 medications: Amikacin, Amorolfine, Bisoprolol, Captopril, Ciprofloxacin, Erythromycin, Ethambutol, Fosfomycin, Isoniazid, Norfloxacin, Sulfamethoxazole, Trimethoprim, Tobramycin, and Valaciclovir.

The alert system and the pharmacists were concordant in the analysis of 4602 (91.93%) prescription lines: 4575 (91.39%) were accepted prescription lines and 27 (0.54%) were the subject of an alert (25 exceeds max daily dose and two under-dose alerts). The alert system and the pharmacists were discordant for 404 (8.07%) prescription lines (Table II).

Kappa coefficient was of 0.106 (CI 95% [0.068 – 0.144]), and thus the agreement between the alert system analysis and the pharmacists' analysis is scored as poor (Table II).

**Table II:** Alert system performance indicators, comparison with the pharmacists' daily analysis of the medication orders

|  |  | Group of pharmacists | | |
| --- | --- | --- | --- | --- |
|  | Alerts | Fired | Not fired | Total |
| The alert system | Fired | 27 | 394 | 421 |
|  | Not fired | 10 | 4575 | 4585 |
|  | Total | 37 | 4969 | 5006 |

**Analysis of cases of concordance between the alert system and the pharmacists**

A senior pharmacist analyzed a random sample of 100 concordant medication orders that is orders for which the pharmacists and the alert system gave the same results. No errors in the checking of these medication orders was identified by the senior pharmacist.

**Analysis of discordance between medication order checks by the alert system and the pharmacists**

Analysis of discordances between the alert system and the group of pharmacists by a senior pharmacist identified three types of discordance (Table III): A- The alert system did not fire an alert and the pharmacists reported an error; B- The alert system fired an over dose alert and the pharmacists do not report an error; C- The alert system fired an under dose alert and the pharmacists do not report an error.

| Type of discordance | Number of prescription lines (N = L + P) | Senior pharmacist analysis | |
|---|---|---|---|
| | | Agrees with the alert system advice (L) | Agrees with the pharmacist advice (P) |
| A- The alert system did not fire an alert and the pharmacists reported an error | N=10 | **L=1** <br> - **For 1** drug prescription line, the pharmacists failed to detect an over dose error | **P=9** <br> - **For 5** drug prescription lines, the clinical decision rule was wrong <br> - **For 1**, the duration of medication was too long and the alert system failed to detect this error <br> - **For 3**, the alert system failed to detect additional prescription lines of the same medication |
| B- The alert system fired an over-dose alert and the pharmacist did not report an error | N= 258 | **L= 201** <br> - **For 146:** the pharmacist failed to detect an over dose associated with an altered eGFR rate <br> - **For 26:** the pharmacist failed to detect an over dose associated with an eGFR rate on the borderline of that authorized <br> - **For 29:** the pharmacist failed to detect an over dose with a normal eGFR rate | **P= 57** <br> - **For 3:** the alert system failed to detect that the dose was adapted to the residual plasma concentration <br> - **For 45:** the alert system failed to detect the weight of the patient <br> - **For 9:** the alert system failed to detect a drug prescription not adapted to the patient blood pressure. |
| C- The alert system fired an under dose alert and the pharmacist does not report an error | N= 136 | **L= 59** <br> - **For 3:** the pharmacist failed to detect an under dose with an eGFR rate on the borderline of that authorized <br> - **For 21:** the pharmacist failed to detect an under dose with an altered eGFR <br> - **For 35:** the pharmacist failed to detect an under dose with a normal eGFR rate | **P=77** <br> - **For 14:** the alert system failed to detect that the dose was adapted to the residual plasmatic concentration <br> - **For 12:** the alert system failed to detect the weight of the patient <br> - **For 6:** the alert system failed to detect that the dose was not adapted to the patients' blood pressure. <br> - **For 45:** the alert system failed to detect that more than one prescription line involved the same medication |
| Total | 404 | 261 | 143 |

**Table III:** Discordance cases between medication orders checked by the alert system and the pharmacists analyzed by a senior pharmacist

The senior pharmacist was in agreement with the alert system advice for 261 of the 404 cases of discordance and with the pharmacists in 143 cases (Table III).

For the 5006 drug prescription lines analyzed (Figure 1), the alert system correctly analyzed 4863 (97.14%) whereas the pharmacists in their daily analysis of these same drug prescription lines correctly analyzed 4745 (94.77%).

According to the senior pharmacist review, the alert system fired 287 true positive alerts, 134 false positives alerts and failed to fire the appropriate alert for nine drug prescription lines (false negatives) (Table IV). The pharmacists fired 36 true positive alerts, one false positive alert and failed to fire the appropriate alert for 260 drug prescription lines (false negatives) (Table III). False negatives cases for both, alert system and pharmacists were rated on our severity scale as without risk or purely preventive.

**Table IV:** Alert system performance indicators, results after a senior pharmacist review of the cases of discordance

|  | Alerts | Senior pharmacist | | |
|---|---|---|---|---|
|  | | Fired | Not fired | Total |
| *The alert system* | Fired | 287 | 134 | 421 |
|  | Not fired | 9 | 4576 | 4585 |
|  | Total | 296 | 4710 | 5006 |

**Impact on clinical decision rules**

Table V describes the clinical decision rules that need to be refined according to the comments of the senior pharmacist.

| | Number of rules per medication | improvements according to the senior pharmacist comments | | | |
|---|---|---|---|---|---|
| | | eGFR rate wrongly configured | Blood pressure not configured | Plasma concentration not configured | Duration of treatment not configured |
| Amikacin | 72 (7,48%) | 0 (0,00%) | 0 (0,00%) | 72 (7,48%) | 0 (0,00%) |
| Atenolol | 30 (3,12%) | 3 (0,31%) | 30 (3,12%) | 0 (0,00%) | 0 (0,00%) |
| Ethambutol | 30 (3,12%) | 0 (0,00%) | 0 (0,00%) | 0 (0,00%) | 0 (0,00%) |
| Metronidazol | 24 (2,49%) | 0 (0,00%) | 0 (0,00%) | 0 (0,00%) | 24 (2,49%) |
| Gentamicin | 40 (4,16%) | 1 (0,10%) | 0 (0,00%) | 40 (4,16%) | 40 (4,16%) |
| Tramadol | 16 (1,66%) | 0 (0,00%) | 0 (0,00%) | 0 (0,00%) | 0 (0,00%) |
| Norfloxacin | 12 (1,25%) | 0 (0,00%) | 0 (0,00%) | 0 (0,00%) | 12 (1,25%) |
| Isoniazid | 30 (3,12%) | 0 (0,00%) | 0 (0,00%) | 0 (0,00%) | 30 (3,12%) |
| Metformin | 18 (1,87%) | 1 (0,10%) | 0 (0,00%) | 0 (0,00%) | 0 (0,00%) |
| tobramycin | 56 (5,82%) | 0 (0,00%) | 0 (0,00%) | 56 (5,82%) | 56 (5,82%) |
| Ramipril | 36 (3,74%) | 0 (0,00%) | 0 (0,00%) | 0 (0,00%) | 0 (0,00%) |
| vancomicin | 36 (3,74%) | 0 (0,00%) | 0 (0,00%) | 36 (3,74%) | 36 (3,74%) |
| Aciclovir | 40 (4,16%) | 0 (0,00%) | 0 (0,00%) | 0 (0,00%) | 0 (0,00%) |
| Allopurinol | 60 (6,24%) | 0 (0,00%) | 0 (0,00%) | 0 (0,00%) | 0 (0,00%) |
| Amoxicillin | 64 (6,65%) | 0 (0,00%) | 0 (0,00%) | 0 (0,00%) | 0 (0,00%) |
| amoxicillin and potassium clavulanate | 136 (14,14%) | 0 (0,00%) | 0 (0,00%) | 0 (0,00%) | 0 (0,00%) |
| Bisoprolol | 34 (3,53%) | 0 (0,00%) | 34 (3,53%) | 0 (0,00%) | 0 (0,00%) |
| Cefotaxim | 32 (3,33%) | 0 (0,00%) | 0 (0,00%) | 0 (0,00%) | 0 (0,00%) |
| ciprofloxacin | 42 (4,37%) | 0 (0,00%) | 0 (0,00%) | 0 (0,00%) | 0 (0,00%) |
| Erythrocine | 18 (1,87%) | 0 (0,00%) | 0 (0,00%) | 0 (0,00%) | 0 (0,00%) |
| fosfomycine | 10 (1,04%) | 0 (0,00%) | 0 (0,00%) | 0 (0,00%) | 0 (0,00%) |
| levofloxacin | 24 (2,49%) | 0 (0,00%) | 0 (0,00%) | 0 (0,00%) | 24 (2,49%) |
| Captopril | 38 (3,95%) | 0 (0,00%) | 0 (0,00%) | 0 (0,00%) | 0 (0,00%) |
| sulfamethoxazole and trimethoprime | 24 (2,49%) | 0 (0,00%) | 0 (0,00%) | 0 (0,00%) | 24 (2,49%) |
| Valaciclovir | 40 (4,16%) | 0 (0,00%) | 0 (0,00%) | 0 (0,00%) | 40 (4,16%) |
| | 962 (100,00%) | 5 (0,52%) | 64 (6,65%) | 204 (21,21%) | 286 (29,73%) |

**Table V:** Clinical decision rules improvements derived according to the senior pharmacist comments and the number of clinical decision rules impacted

The initial specifications developed in collaboration with the pharmacists included the implementation of 962 rules. Only five (0.52%) of these rules did not fulfill the expected requirements because of incorrect configuration of the eGFR threshold (Table V). The other cases are related to various issues and can be used for improvements as discussed in the following section.

## 5. Discussion

We implemented a rule-based clinical decision alert system to support drug dosing prescription adjustment. We compared its performance over seven months with routine daily medication order analysis by several pharmacists trained to this task. The alert system and the pharmacists showed a relatively poor inter-rater agreement; the alert system fired more appropriate alerts than the pharmacists and made fewer errors than pharmacists in analyzing drug dose prescriptions: 143 errors for the alert system versus 261 for the pharmacists. Most of diagnostic errors made by the pharmacists, but not the alert system, were 'false negative', i.e. errors not detected by the pharmacists corresponded to drug prescription lines with an exceeded maximum daily dose or under dosing. This is important in routine clinical practice because it is important that the system detects all potential errors. In addition, during the study period, the pharmacists were not able to analyze a significant number of drug prescription lines, 2097 (25.42%), because of understaffing; this issue obviously does not affect the alert system (Pharmacist's workload issue can be also addressed by adapting the alert system to the prescriber physicians). The alert system failed to fire an alert for nine drug prescription lines (table 3); these cases are discussed in the following as potential alert system improvements.

The design of a system based on clinical decision rules involves three main steps: creation or developing the rules; testing and validating the rules; and finally, analyzing the effects of the rules on clinician behavior or patient outcome (17). This study represents the second of these steps for a set of clinical decision rules designed and developed at the HEGP since 2008 (10)(11). In this set of rules, only five (0.52%) rules did not fulfill the expected requirements (Table V). We are convinced that this small number of "wrong" rules is related to the use of an "in-silico" retrospective testing process (15). This design step (15) allows to established a 'partial' external validity of these rules. The study we report here completes the validation of these rules. The next step is thus an impact analysis study.

The last step of our method was the expert senior pharmacist review of both CDSS alert interventions and the pharmacist's interventions. Post expert reviews of CDSS alerts response have already been used in other studies to validate clinical alerts (18)(19)(20). McCoy et al. (20) proposed a framework based on experts reviews to judge clinical alerts appropriateness. Despite the fact that the senior expert review was done in optimum conditions by comparison with the daily pharmacist's analysis of medication orders; a limitation in the method we report here was the use of only one expert review.

According to the analysis of the senior pharmacist (table V), three important improvements related to the clinical decision rules should be considered:

- The first group concerns rules for which clinical conditions are missing. Clinical decision rules that check orders of: atenolol, bisoprolol, captopril and ramipril, should be configured to include the blood pressure levels at time of ordering as a parameter. On the initiation of treatment, the drug doses should be adapted to the patient eGFR. Subsequently, however, drug doses should be adapted to the clinical response, in this case, according to the blood pressure. Even if the initial drug dosage prescribed respects the eGFR adaptation rules, the subsequent dose adjustment should be based on patient clinical response such as orthostatic hypotension or persistence of high blood pressure.

- The second group concerns clinical decision rules for which duration of treatment and/or cumulative dosage are of importance. Clinical decision rules that check orders of: antibiotics (amikacin, metronidazole, gentamicin, norfloxacin, isoniazid, tobramycin, vancomycin, levofloxacin, sulfamethoxazole/trimethoprim and valaciclovir), should be configured to include the duration of treatment as a parameter. Indeed, drug prescription of antibiotics agents for a long period of treatment can lead to the emergence of resistance. Recommendations of treatment duration of anti-infectious agents depend on the type of infectious disease that has be diagnosed. For example, for Levofloxacin, the treatment duration should be of 7 days in the case of cystitis and 3 to 6 weeks in the case of a prostatitis.

- The third group concerns clinical decision rules for which individual response and plasma concentration measurements can be required. Clinical decision rules that check orders of: amikacin, gentamicin, tobramycin, vancomycin, valaciclovir and digoxin should be configured to include the plasma concentration as a parameter on which alerts may be based. For example: after 24 hours of a first prescription of Amikacin to a patient, even if the drug dose is adapted to the patient eGFR, the patient residual plasma concentration should be checked and should be lower than or equal to 4 mg/L.

Several of the reasons that led to the failure of the alert system were associated with the uncertainty of medical knowledge. Uncertainty in handling medical knowledge can be of several types (21):

- Lack of information: the data necessary to fire an alert are not always available, or may be available but unreliable, or not sufficiently structured to be processed by an electronic device. This was the case for drug prescription lines where the weight of the patient was not available or was only reported in a Word file attached to the patient EHR.

- Fuzziness in determination of which clinical data fire an alert: this was the case for drug prescription lines where alerts were fired because of the patient eGFR rate but the senior pharmacist indicated that the dose was, appropriately, adapted to the residual plasmatic concentration or to the patient blood pressure or to the treatment duration, and not to the eGFR of the patient.

- Vagueness in the formulation of certain recommendations on which we based the implementation of some of the clinical decision rules: the exact meaning of "the expert panel prefer…" or "…in such cases kalemia must be checked closely", for example, is unclear.

Issues related to the lack of structured data can be partially covered by NLP tools. Different architectures can be implemented: NLP systems can be integrated within the alert system or coupled with it to various degrees of agility; NLP systems can be governed by the alert system or implement knowledge and logic necessary to support decisions; some NLP systems have been developed for specific tasks or alternatively, a generalized tool can be customized for particular tasks (22).

Issues related to the vagueness of clinical knowledge can be approached by extending the system to fuzzy logic. There are few examples of the practical application of this technique in clinical decision support (23); it is nevertheless a very promising avenue of research for optimizing our clinical decision rules-based alert system (24).

The alert system fired 287 appropriate alerts, these alerts were rated by the senior pharmacist as without risk or purely preventive according to our severity scale. Even if the senior pharmacist judged that firing these alerts was necessary, we think that in order to avoid the alert fatigue syndrome in daily practice, alerts without risk or purely preventive alerts should be implemented in non-interruptive mode and targeted both towards the physicians and the pharmacists. Alerts related to contraindications will be implemented in interruptive mode (25). Both interruptive and non-interruptive alerts can be useful in daily clinical practice (26), although there is some difficulty in determining the threshold between a non-interruptive alert and an interruptive alert. More tests should be performed using our approach reported in (15) to define the different parameters (the drug dosage, the patient eGFR,…) and the values of each parameters which can switch an alert from an non-interruptive mode to an interruptive mode.

Discrepancies between pharmacists in daily medication orders analysis may partly explain the low concordance rate between the group of pharmacists and the alert system. Checklists seem to be a key instrument to reduce individual variations between staff members doing the same tasks, to reduce human error and ensure that quality standards are maintained (27)(28). For pharmaceutical analysis of drug orders, checklists could be a list of clinical action items related to one or several clinical parameters arranged in a systematic and easy manner allowing the pharmacist to remember and / or to record the presence / absence of each item listed and to ensure that all have been checked.  This checklist should be integrated into the CPOE/EMR system in order to be used by pharmacists during their daily analysis of drug orders (29).

**Conclusion**

We have been investigating individual variation between pharmacists analyzing physician medication orders. We observed that more severe the renal function impairment was associated with a greater number of unsatisfactory drug prescription lines in medication orders and with a better pharmacist analysis of that type of medication order. In other words, the pharmacists' diagnostic performance was better for more dangerous medication orders than for less dangerous orders (manuscript in preparation). These results and the results of the study we report here reinforce our conviction that the computerized alert system is complementary to pharmacists for their contributions to drug prescription safety. Knowing that a large number of drug prescription lines are not been analyzed by pharmacists because of understaffing, using the alert system with a non-interruptive mode for some alerts could relieve pharmacists of some of the routine work. Consequently, they could focus on detecting more 'dangerous' medication orders that would have passed the first barrier of focused physicians' targeted alerts and thereby enhance the safety of the process of drug prescription.

**Funding**

This work was supported by Programme de recherche en qualité hospitalière (PREQHOS-PHRQ 1034 SADPM), The French Ministry of Health, grant number 115189.